\documentclass[review]{elsarticle}
\usepackage{lineno,hyperref}
\modulolinenumbers[5]
\usepackage{amsfonts}
\usepackage{amsmath}
\usepackage{amssymb}
\usepackage{bm}
\usepackage{booktabs}
\usepackage{dsfont}
\usepackage{graphicx}
\usepackage{epstopdf}
\usepackage{epsfig}
\usepackage{url}
\usepackage{subfigure}
\usepackage{color}
\usepackage{algorithmic}
\usepackage{algorithm}
\usepackage{threeparttable}
\usepackage{lineno}
\usepackage{multirow}
\usepackage{ragged2e}
\usepackage{diagbox}
\usepackage{float}
\usepackage{setspace}
\usepackage{stfloats}
\usepackage{mathabx}
\usepackage{amsthm}
\usepackage{amsopn}

\graphicspath{{./figure/}}

\renewcommand{\raggedright}{\leftskip=0pt \rightskip=0pt plus 0cm}

\newsavebox\CBox

\makeatletter
\renewcommand{\maketag@@@}[1]{\hbox{\m@th\normalsize\normalfont#1}}%
\makeatother

\journal{Journal}

\begin{document}

 \begin{frontmatter}

\title{Fair Streaming Feature Selection }
\author[address1]{Zhangling Duan}
\ead{duanzl1024@ahu.edu.cn}

\author[address1]{Tianci Li}
\ead{y23301058@stu.ahu.edu.cn}

\author[address2]{Xingyu Wu\corref{mycorrespondingauthor}}
\ead{xingy.wu@polyu.edu.hk}

\author[address3]{Zhaolong Ling}
\ead{zlling@ahu.edu.cn}

\author[address1]{Jingye Yang}
\ead{y23301044@stu.ahu.edu.cn}

\author[address1]{Zhaohong Jia\corref{mycorrespondingauthor}}
\ead{zhjia@mail.ustc.edu.cn}

\address[address1]{School of Internet, Anhui University, Hefei 230601, China.}
\address[address2]{Department of Computing, The Hong Kong Polytechnic University, Hong Kong SAR, China, 999077}
\address[address3]{School of Computer Science and Technology, Anhui University, Hefei 230601, China.}

\cortext[mycorrespondingauthor]{Corresponding author}
%

\begin{abstract}
Streaming feature selection techniques have become essential in processing real-time data streams, as they facilitate the identification of the most relevant attributes from continuously updating information. Despite their performance, current algorithms to streaming feature selection frequently fall short in managing biases and avoiding discrimination that could be perpetuated by sensitive attributes, potentially leading to unfair outcomes in the resulting models. To address this issue, we propose FairSFS, a novel algorithm for \underline{Fair} \underline{S}treaming \underline{F}eature \underline{S}election, to uphold fairness in the feature selection process without compromising the ability to handle data in an online manner. FairSFS adapts to incoming feature vectors by dynamically adjusting the feature set and discerns the correlations between classification attributes and sensitive attributes from this revised set, thereby forestalling the propagation of sensitive data. Empirical evaluations show that FairSFS not only maintains accuracy that is on par with leading streaming feature selection methods and existing fair feature techniques but also significantly improves fairness metrics.

\end{abstract}

\begin{keyword}
Fair Feature Selection, Streaming features,  Markov blanket.
\end{keyword}

\end{frontmatter}

\section{Introduction}
With the arrival of the big data era, the continuous emergence of new features in data streams presents unprecedented challenges~\cite{perkins2003online,glocer2005online}. The feature space is no longer static but evolves over time~\cite{li2017feature,wu2012online}. In a scenario where a social media platform utilizes streaming feature selection to determine the content delivered to users~\cite{wu2012online}, the platform aims to select the most relevant material based on individual interests and preferences, thereby providing a personalized user experience~\cite{hosu2020koniq,li2017feature}. This necessitates algorithms that can dynamically update the feature set as new data arrives in real-time~\cite{wu2012online}, ensuring that the model always predicts based on the latest relevant information.\par

In recent years, researchers propose various stream feature selection algorithms, such as OFS~\cite{perkins2003online}, OSFS~\cite{wu2012online}, and SAOLA~\cite{yu2016scalable}, which can dynamically update the feature set in real-time, ensuring that the model is always based on the latest relevant information for prediction. This method has significant advantages in dealing with stream feature selection problems~\cite{zhou2020feature,zhou2022online}. Nevertheless, in a dynamic stream feature environment, traditional stream feature selection algorithms that only seek high-correlation features are no longer sufficient to address the challenge of fairness~\cite{belitz2021automating}. We must ensure that the selected features do not lead to unfair decisions against certain groups, maintaining the fairness and adaptability of the model~\cite{corbett2017algorithmic}.\par

 In data science and machine learning, fairness is particularly concerned with avoiding the unfair impact of algorithms and models on specific groups or individuals during the decision-making process~\cite{galhotra2022causal,mehrabi2021survey}. It ensures that the model does not discriminate against or exhibit an unfair bias towards specific individuals or groups in areas such\begin{figure}[!htbp]
\centering
 \includegraphics[width=4.6in, height=1.5in]{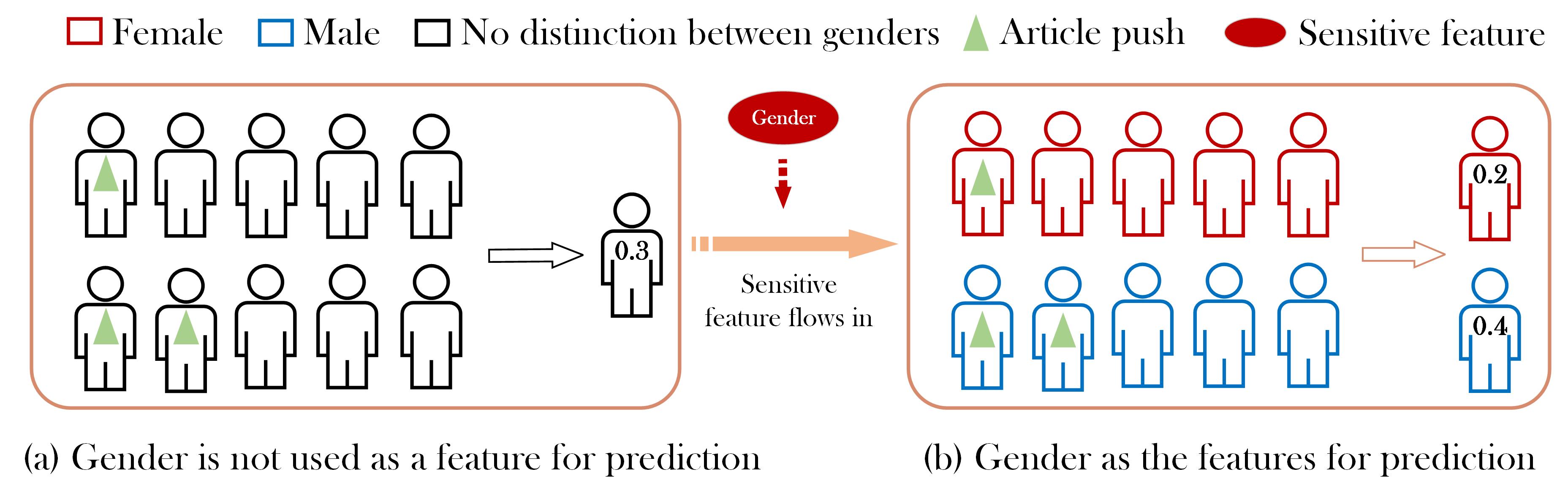}
 \caption{The use of gender features in the content recommendation process may lead to bias and discrimination.}
\end{figure} as credit assessment~\cite{bernanke1988credit}, justice~\cite{sutera1993history}, and medicine~\cite{pellegrino1966medicine} based on sensitive attributes like race, gender, or age. Ensuring the fairness of the model has become a critical focus~\cite{grgic2016case}.\par

However, in real-world applications, we may encounter challenges related to fairness in stream feature selection~\cite{corbett2023measure,grgic2016case}. Consider, for instance, a social media platform where the content recommendation system is trained to prioritize articles related to work and technology for male users, while predominantly recommending articles about beauty and fashion for female users. As illustrated in Figure 1, the left panel (a) depicts the proportion of work- and technology-related articles recommended to a user before the gender-sensitive feature is introduced, which is 0.3. Upon the incorporation of the gender-sensitive feature, the right panel (b) reveals that the model's recommendations are influenced by this feature. With gender as the evaluation criterion, the proportion of work- and technology-related articles recommended to male users increases to 0.4, while for female users, it decreases to 0.2. This approach unquestionably introduces gender bias in content recommendations, potentially reducing the opportunities for female users to access technology-related information crucial for their career development, thereby potentially limiting their growth and advancement in the tech field.\par

To address this issue, we begin to explore how to incorporate fairness principles into the stream feature selection process. This requires the model to avoid biases based on sensitive attributes such as race, gender, or age.
Given this, we need an algorithm capable of handling sensitive features within the streaming feature environment to ensure the model does not unfairly treat any group during decision-making. This poses a significant challenge for fair feature selection in the context of streaming features.
To address this challenge, the main work and contributions of this paper are as follows:

\begin{itemize}
    \item We introduce the problem of fair feature selection in a streaming data environment. Furthermore, we examine the difficulties in achieving a fair feature set for this problem from a theoretical standpoint.

    \item We propose a novel fair streaming feature selection algorithm, FairSFS, which can dynamically update the feature set in real-time as features continuously flow in, and based on the real-time feature set, it identifies the correlations between classification variables and sensitive variables, effectively blocking the flow of sensitive information.
    \item In experiments conducted on seven real-world datasets, FairSFS not only matches the accuracy of three streaming feature selection algorithms and two fair feature selection algorithms but also effectively achieves fair feature selection in a streaming data environment.
\end{itemize}

The remainder of this paper is structured as follows: Section 2 reviews the related work in the domain of feature selection. Section 3 introduces the fundamental definitions. Section 4 delineates the FairSFS algorithm, provides a proof of correctness for the algorithm, and offers an in-depth analysis. Section 5 presents the empirical outcomes and the associated examination. Finally, we summarize our findings in Section 6 and propose directions for subsequent investigation.

\section{Related work}
This paper aims to address the problem of fairness deficits in streaming feature selection algorithms when dealing with data involving sensitive features, by applying the principles of fair feature selection. Therefore, this section presents work related to streaming feature selection algorithms and fair feature selection algorithms.\par
\subsection{Streaming feature selection}\par
Several research efforts have been directed towards tackling the challenges associated with streaming features. Perkins and Theiler~\cite{perkins2003grafting} addressed the issue of streaming feature selection and introduced the Grafting algorithm, which is a staged gradient descent method designed for streaming feature selection. Grafting conceptualizes the selection of relevant features as an essential component of the predictor learning process within a regularization-based learning architecture. It employs two iterative steps to optimize an L1-regularized maximum likelihood: the optimization of all free parameters and the selection of new features. Grafting operates incrementally, incrementally constructing the feature set while concurrently training the prediction model via gradient descent. In each iteration, Grafting employs a rapid, gradient-based heuristic method to pinpoint the features that are likely to improve the current model, and it consecutively refines the model using gradient descent. Expanding on this methodology, Glocer et al.~\cite{glocer2005online} adapted the algorithm to tackle edge detection issues in grayscale imagery.
 Although Grafting is capable of managing streaming features, it necessitates the pre-selection of regularization parameter values, which dictates which features are most probable to be chosen in each iteration. The requirement for suitable regularization parameters inherently demands knowledge about the global feature set. Consequently, Grafting may not perform optimally when dealing with streaming features of unknown dimensions.

Ungar et al. and Zhou et al. delved into the realm of streaming feature selection and introduced two innovative algorithms, information-investing and Alpha-investing~\cite{ungar2005streaming,zhou2006streamwise}, which are grounded in the principles of streaming regression. Dhillon et al. further advanced the Alpha-investing approach by introducing a multi-stream feature selection algorithm capable of managing multiple feature classes simultaneously~\cite{dhillon2010feature}. Alpha-investing conceptualizes the set of candidate features as a stream that is generated dynamically, with new features being sequentially evaluated for inclusion in the predictive model. Alpha-investing excels in managing candidate feature sets of unknown or potentially infinite scope. It adjusts the threshold for error decrement, necessary for the inclusion of novel features in the predictive model, by utilizing either linear or logistic regression in an adaptive manner. However, a significant limitation of Alpha-investing is its sole focus on feature addition without subsequent evaluation of the redundancy among the selected features once new ones have been integrated.\par
Although  streaming feature selection algorithms excel in adapting to dynamic changes in data features and can timely update the selected features, they might overlook potential biases introduced in model predictions, especially when involving sensitive features. This indicates that while these algorithms have significant advantages in data processing speed and adaptability, they fall short in ensuring the fairness of decisions.\par
\subsection{Fair machine learning}\par
The pursuit of fairness in machine learning algorithms has become a critical domain of research~\cite{pessach2022review,yu2019multi}, aimed at mitigating biases and disparities inherent in these models. There is a growing acknowledgment of the importance of fairness in preserving human rights, ethical standards, and social equity. Achieving balanced results for various populations is crucial for enhancing the credibility of technological infrastructure and for promoting fair societal advancement.
 Numerous recent studies have introduced methods to enhance the fairness of machine learning models, generally classified into three categories: pre-processing, in-processing, and post-processing~\cite{pessach2022review}.
Pre-processing techniques primarily entail modifying the training data prior to its input into machine learning algorithms. Early pre-processing methods, such as those proposed by Kamiran and Calders~\cite{kamiran2012data} and Luong et al.~\cite{luong2011k}, involve altering labels or reweighting specific instances to achieve fairer classification results. Typically, samples proximate to the decision boundary are more susceptible to label changes, as they are most likely to be misclassified. Contemporary approaches suggest altering data feature representations to render subsequent classifiers more equitable.
In-processing methods involve direct modifications to machine learning algorithms~\cite{zafar2017fairness}, with an emphasis on integrating fairness considerations during the training phase. For instance, Zafar et al.~\cite{zafar2017fairness} and Woodworth et al.~\cite{woodworth2017learning} propose incorporating fairness constraints into classification models to satisfy equalized odds or other impact-related metrics. Bechavod and Ligett~\cite{bechavod2017penalizing} suggest including fairness penalties within the objective function to enforce metrics such as false positive rate (FPR) and false negative rate (FNR). Zemel et al.~\cite{zemel2013learning} combined fair representation learning with procedural models by employing a logistic regression-based multi-objective loss function, while Louizos et al.~\cite{louizos2015variational} apply this concept through the use of a variational autoencoder.
Post-processing techniques primarily focus on adjusting the output scores of classifiers to render decisions fairer. For example,  Corbett-Davies et al\cite{corbett2017algorithmic} and Menon and Williamson~\cite{menon2018cost} propose establishing distinct thresholds for each group, aiming to maximize accuracy while reducing differences at the population level.
In the domain of Graph Neural Networks (GNNs), Zhang et al.~\cite{zhang2023fpgnn}  introduce a novel deep model, FPGNN (Fair Path Graph Neural Network), crafted to curtail the spread of sensitive data within GNN models. Utilizing a scalable random walk technique (termed $"$fair path$"$), it identifies higher-order nodes that play a crucial role in maintaining fairness at the node level.Nevertheless, this method might lead to the neglect of sensitive characteristics connected to nodes with low correlation and an overemphasis on the influence of sensitive nodes with high correlation on their neighboring nodes. To rectify these issues, the SRGNN (Strategic Random Walk Graph Neural Network) algorithm~\cite{zhang2024learning} has been introduced. SRGNN takes into account both low-degree and high-degree nodes within GNN models, considering their effects on fairness in representation during the decision-making phase.\par
Overall, while current fairness-enhancing algorithms have put forward numerous equitable approaches, there remains a gap in their ability to effectively manage features within dynamically evolving data streams.
 Building on the work above, this paper attempts to combine fair feature selection algorithms with  streaming feature selection algorithms, proposing a fair feature selection algorithm in a streaming data environment.
\section{Definitions}
In this section, we will delve into streaming feature selection and fairness, exploring the definitions and theorems of streaming feature selection and fairness.\par

\textbf{ (Streaming Features)~\cite{wu2012online}:} Streaming features are features within the feature space that evolve over time, while the training data's sample space remains fixed. These features are introduced sequentially, one by one, or continuously generated.\par

The distinctiveness of feature selection in streaming features, as opposed to traditional selection methods, lies in:
(1) The dynamic and uncertain nature of the feature space, where dimensions may continually increase, potentially becoming infinite.
(2) The streaming aspect of the feature space, where features arrive sequentially, and each new feature is promptly processed upon arrival.\par

\textbf{ (Conditional Independence)~\cite{pearl1988probabilistic}:} If variables \(X\) and \(Y\) are conditionally independent given \(S\), then \(P(X,Y|S) = P(X|S)P(Y|S)\), denoted as \(X \!\perp\!\!\!\perp Y|S\).\par

\textbf{ (D-separation)~\cite{pearl1988probabilistic}:} For variables \(X, Y \in U\) and a set \(S \subseteq U \setminus \{X, Y \}\), a path \(\pi\) between \(X\) and \(Y\) given \(S\) is blocked if and only if (1) the non-colliders on \(\pi\) are in \(S\), or (2) \(S\) lacks all colliders on \(\pi\) or their descendants. If \(S\) blocks all paths between \(X\) and \(Y\), then \(X\) and \(Y\) are D-separated by \(S\).\par

\textbf{(Faithfulness)~\cite{spirtes2001causation}:} A $BN < V, G, P >$ is faithful iff all conditional dependencies between features in $G$ are captured by $P$.
Faithfulness indicates that in a BN, $X$, and $Y$ are independently conditioned on a set $S$ in $P$ iff they are d-separated by $S$ in $G$.

\textbf{ (K-fair)~\cite{salimi2019interventional}:} Fix a set of attributes \(K \subseteq V-\{S, O\}\). An algorithm \(\ell:\) Dom(X) \(\rightarrow\) Dom(O) is \(K\)-fair w.r.t. a sensitive attribute \(S\) if for any context \(K = k\) and outcome \(O = o\), the following holds:
\begin{equation}
 Pr(O = o \mid do(S = 0), do(K = k)) = Pr(O = o \mid do(S = 1), do(K = k))
\end{equation}
If the algorithm is K-fair for every set \(K\), it is said to be fair by intervention. Additionally, in the intervention graph \(G'\) (with incoming edges from \(S\) to \(K\) removed), the sensitive attribute \(S\) is unrelated to \(Y'\) under \(K\), i.e., \(S\) and \(Y'\) are separated by \(K\) in graph \(G'\).\par
Considering dataset \(D\), where \(V = S \cup X \cup Y\) contains the sensitive variable \(S\), the non-sensitive variable set \(X = \{X_1, X_2, \ldots, X_n\}\), and the label variable \(Y\). \(MB_Y\) and \(MB_S\) are the Markov Blanket variable sets of \(Y\) and \(S\) respectively, including the children and spouses of \(Y\) and \(S\). \(Y'\) is the target variable obtained after training from the subset \(T \subseteq V\). The definition of fairness is as follows:

Given dataset \(D\), for the sensitive variable \(S\) and non-sensitive variable set \(X\), if the target variable \(Y'\) trained from the subset \(T \subseteq V\) satisfies a specific fairness property, then \(T\) is considered to have fair features.\par

\textbf{(Do Operator)~\cite{pearl2009causality}:} Intervention on attribute \(X\), denoted as \(X \leftarrow x\), is effectively implemented by assigning the value \(x\) to variable \(X\) in the modified causal graph \(G'\), where \(G'\) is the same as \(G\) except for the elimination of all incoming edges to \(X\).\par

The Do operator is consistent with the graphical interpretation of interventions. Specifically, an intervention denoted as \(do(X) = x\) is equivalent to conditioning on \(X = x\) when \(X\) has no ancestors in graph \(G\).\par

\textbf{ (Markov Blanket)~\cite{pearl1988probabilistic}:} In a faithful Bayesian network, each variable has only one Markov blanket (MB) consisting of its parents, children, and spouses (parents of its children).
Given the MB of \(T\), denoted \(MB_{T}\), all other variables are conditionally independent of \(T\).
\begin{equation}
 X \!\perp\!\!\!\perp T\mid MB_{T}, \forall  X \in V \backslash MB_{T} \backslash \{X\}
\end{equation}
Pearl introduced interventions, involving altering the state of an attribute to a specific value and observing the effects.

\textbf{ (Fair Features):} A feature set \(T\) is deemed fair if: (1) The classifier trained on \(T\) meets K-fairness criteria for the predictive variable \(Y'\); (2) The features in \(T\) adequately represent the class variable \(Y\).\par

Here \(T \subseteq V\). Feature selection aims to identify a fair subset \(T\) by understanding relationships among features, the class variable, and sensitive variables. The objective is to ensure that the target variable \(Y'\) trained using these features meets fairness criteria.

\section{Fair streaming feature selection}

Due to the limitations of traditional fair feature selection algorithms in handling streaming data scenarios, we propose a fair streaming feature selection algorithm—FairSFS. In this section, we first introduce the FOFS algorithm and progressively verify its theoretical correctness in Section 4.1, then analyze FairSFS through examples in Section 4.2.

\subsection{ Algorithm implementation}

During the feature selection process, the FairSFS algorithm streams features one by one. Each incoming feature is evaluated for its independence from other selected features through conditional independence tests to ensure that the selected features meet fairness requirements.

Our goal is to use streaming features as inputs (denoted by $X_i$ for the $i$-th feature) to identify features that are relevant to a specific target variable $T$ and block paths with a sensitive feature $S$. This process involves two main steps aimed at dynamically identifying a set of features that are related to the target variable and unaffected by the sensitive feature. The detailed steps are as follows:\par
\begin{algorithm}[t]
\caption{FairSFS algorithm}
\label{alg:your_algorithm}
\setstretch{1.15}
\ \ \textbf{Input}: $D$: dataset; $T$: the target; $S$: sensitive feature\par
\ \ \textbf{Output}: $MBT$: Markov blanket of $T$
\begin{algorithmic}[1]
\STATE $MB_S \leftarrow \emptyset; MB_T \leftarrow \emptyset;$
\REPEAT
\STATE /* Step 1: Preliminary classification of $X$ */
\STATE $X \leftarrow \text{get a new feature};$
\IF{$\text{dep}(S, X | MB_S)$}
\STATE $MB_S \leftarrow MB_S \cup \{X\};$
\ELSIF{$\text{dep}(T, X | MB_T)$}
\STATE $MB_T \leftarrow MB_T \cup \{X\};$
\ENDIF
\STATE /* Step 2: Select features from $MB_S$'s spouses that belong to $MB_T$ */
\FOR{each $A \in MB_S$}
\IF{$\text{Ind}(S, A | \emptyset)$ \AND $\text{dep}(T, A | MB_T)$}
\STATE $MB_T \leftarrow MB_T \cup \{A\};$
\ENDIF
\ENDFOR
\UNTIL{condition}
\STATE \textbf{return} $MB_T;$
\end{algorithmic}
\end{algorithm}

Step 1: Initially, we sequentially input features from the dataset. For each newly arriving feature $X_i$, we perform a preliminary classification. We check whether there is a dependency relationship between $X_i$ and the sensitive feature $S$ under the condition of the already selected sensitive feature set $MB_S$; if there is a dependency, i.e., $X_i$ is not independent of $S$, then we add this feature $X_i$ to the $MB_S$ set. If $X_i$ is independent of $S$, according to Lemma 1, $X_i \perp\!\!\!\perp S \mid MB_{S}(i)$, it can be considered fair under the context of the sensitive attribute $S$ when $X_i$ is input. On this basis of fairness, we further check whether $X_i$ is dependent on the target variable $T$ under the condition of the target variable set $MB_T$; if there is a dependency, indicating that $X_i$ and $T$ are not independent, and according to Lemma 2, the fair feature set $MB_T$, after adding a fair causal feature $X_i$, $MB_T'$ is still a solution to the fair feature selection problem, therefore, we add $X_i$ to the $MB_T$ set.

\textbf{Lemma 1: }\textit{At the time of feature $X_i$ input, if the features in $M_T$ are conditionally independent of $S$ given the Markov boundary $MB_S(i)$ of $S$, i.e., $Z \perp\!\!\!\perp S \mid MB_{S}(i)$ ($Z \in M_T$), then the features in $M_T$ at the time of $X_i$'s input can be considered fair concerning the sensitive attribute $S$.}

 \renewcommand\qedsymbol{\ensuremath{\blacksquare}}
\begin{proof}
 Here, we use $MB_{S}(i)$ to denote the state of the Markov blanket of $S$ when the $i$-th feature, $X_i$, is input. Given the condition $Z \perp\!\!\!\perp S \mid MB_{S}(i)$ ($Z \in M_T$), it implies that the features in $M_T$ do not capture any information about the sensitive variable $S$. Therefore, all paths from $S$ to the target $T'$ through $M_T$ are blocked. Mathematically, we derive:
\begin{equation}
\begin{aligned}
&Pr[T'| \text{do}(S), MB_{S}(i)] \\
&= \Sigma_{M_{T}} Pr[T'| M_{T}, \text{do}(S), MB_{S}(i)] \cdot Pr[M_{T} | \text{do}(S), MB_{S}(i)]\\
1)
&= \Sigma_{M_{T}} Pr[T'| M_{T}, \text{do}(S), MB_{S}(i)] \cdot Pr[M_{T} |  MB_{S}(i)]\\
2)
&=Pr[T'| MB_{S}(i)]\\
\end{aligned}
\end{equation}\par
1) Since $Z \perp\!\!\!\perp S \mid MB_{S}(i)$ ($Z\in M_{T}$), which means that the features in $M_{T}$ are conditionally independent of $S$ given $MB_{S}(i)$, it indicates that all dependent paths from $M_{T}$ to $S$ are blocked by $MB_{S}(i)$. Therefore, a classifier trained using $M_{T}$ will not capture any sensitive information about $S$, because the sensitive information cannot be transmitted through $M_{T}$ given $MB_{S}(i)$. Additionally, performing the $\text{do}(S)$ operation, which is equivalent to removing all incoming edges from $S$ to other nodes in the causal graph, thus cutting off the influence of $S$ on $T$.

2) Assume that $T$ depends only on the variables in $M_{T}$ under all circumstances. Given $M_{T}$, $T$ is conditionally independent of $S$. Therefore, even performing the $\text{do}(S)$ operation does not change the conditional distribution of $T$, since the distribution of $T$ is mediated only through $M_{T}$. After performing the $\text{do}(S)$ operation, as all incoming edges to $S$ are removed, there is no longer any direct or indirect connection between $T$ and $S$, ensuring that $\Pr[T' \mid M_{T}, \text{do}(S), MB_{S}(i)] = \Pr[T' \mid M_{T}, MB_{S}(i)]$.
\end{proof}
\textbf{Lemma 2: }\textit{If a set of fair causal features $D$, after adding a fair causal feature $X$, results in $D'$ which is still a solution to the fair causal feature selection problem, then the classifier trained on $D'$ is also causally fair.}

\begin{proof}
   Considering \(T'\) as the predicted outcome of the classifier trained on dataset \(D\), we envision \(G'\) as a revised causal network where all the directed links towards \(S\) have been severed. According to the principle of causal fairness, any path originating from the sensitive attribute \(S\) and leading to \(T'\) is blocked in \(G'\) (i.e., \(S\) is conditionally independent of \(T'\) given \(G'\)). Since \(T'\) is a dependent variable in \(D\), any paths from \(S\) to the parents of \(T'\) in \(D\) are also obstructed (i.e., \(S\) is conditionally independent of the parents of \(T'\) given \(G'\)). Consequently, we posit:
\begin{equation}
\begin{aligned}
Pr [T' \mid do(S)=s, MB_{S} ]
&= \Sigma_{pa(T')=c} Pr[T'=y \mid pa(T')=c, MB_{S}]\\ &\cdot Pr[pa(T')=c \mid do(S)=s, MB_{S}]\\
\end{aligned}
\end{equation}

Because performing a do-operation on $S$ is equivalent to creating a new causal graph $G'$, where the value of $S$ is set to $s$, and $T'$ has only the parent nodes $D'$, thus $Pr[pa(T')=c \mid do(S)=s, MB_{S}] = Pr_{G'} [pa(T')=c \mid S=s, MB_{S} ]$.
Since $T'$ is trained over $D'$, $pa(T') \subseteq D'$, $ S \!\perp\!\!\!\perp pa(T') \mid G'$, $Pr_{G'}$[$pa(T')=c \mid$ S=s, $MB_{S}$]=$Pr_{G'}$[$pa(T')=c \mid MB_{S}$]. In $G'$, $Pr_{G'} [pa(T')=c \mid S=s, MB_{S}] = Pr_{G'} [pa(T')=c \mid MB_{S}]$, $T'$ satisfies Definition 5, therefore, $D'$ is causally fair.
\end{proof}
Step 2: In Step 1, we have identified features that belong to $MB_T \setminus MB_S$. According to Lemma 3, the continuous inflow of features affects the fairness of previous features; hence, we search within all$'$spouses$'$ in $MB_S$ for features that were incorrectly assigned to $MB_S$ before the inflow of feature $X_i$. We first check for features $A$ that are independent of $S$ without any other conditions. We then further check whether $A$ is dependent on $T$ given $MB_T$. If $A$ depends on $T$, we add $A$ to $MB_T$. This process is repeated until all features in $MB_S$ have been considered. Ultimately, we return $MB_T$, which represents the Markov blanket of the target variable $T$.

\textbf{Lemma 3: }\textit{If  nodes $X$ in the spouses of $MB_{S}$ satisfying
\( X \not\perp\!\!\!\perp S | M(j) \), where $M(j) \subseteq MB_{S}(j)$, but with the arrival of the $k$-th feature, for some $M(k) \subseteq MB_{S}(k)$, \( X \perp\!\!\!\perp S | M(k) \), then $X$ is also causally fair.}
 \renewcommand\qedsymbol{\ensuremath{\blacksquare}}
\begin{proof}
The continuous influx of features may impact the fairness of prior features. For example, upon arrival of the \(j\)-th feature, \( X \not\perp\!\!\!\perp S | MB_{S}(j) \), but upon arrival of the \(k\)-th feature (\(k>j\)), \( MB_{S}(k) \), \( X \perp\!\!\!\perp S | MB_{S}(k) \). According to Lemma 2, under the condition \(MB_{S}(k)\), feature \(X\) does not contain any information about the sensitive attribute \(S\), thus all paths from \(S\) to the target \(Y'\) through \(X\) are blocked. Since all paths from \(X\) to \(S\) are blocked upon arrival of the \(k\)-th feature, a classifier trained using \(X\) will not capture any sensitive information about \(S\), and \( X \perp\!\!\!\perp S | M(k) \) holds.
\begin{equation}
\begin{aligned}
&Pr[T'| \text{do}(S), M(k)]= \Sigma_{X} Pr[T'| X, M(k)] \cdot Pr[X | \text{do}(S), M(k)]
\end{aligned}
\end{equation}
The variable $T'$ only depends on the feature $X$ in the environment before the arrival of the $k$-th feature. Given $M(k)$, $T'$ is independent of $S$.Furthermore, the node \( S \) does not receive any incoming edges. Consequently, by applying the do-calculus rule, we can deduce that \( T' \) is independent of \( S \) in the modified graph where the incoming edges to the \( S \) node have been eliminated, $Pr[T'|X, \text{do}(S), M(k)] = Pr[T'|X, M(k)]$, and $X$ is a fair causal feature. Hence, in Step 2, we need to search for such features from the spouses of $S$.
\end{proof}
\subsection{Algorithm analysis}\par
In this section, we'll outline the specific goals of the FairSFS algorithm for feature selection using an illustrative example. Current methods depend on manually chosen acceptable variables to ensure fairness, but this approach lacks clear standards, leading to unreliable results and possibly irrelevant features. Additionally, existing fair feature selection algorithms lack real-time and dynamic capabilities needed in today's data stream environments.\par
\begin{figure}[!htbp]
\centering
 \includegraphics[width=4.8in, height=2.0in]{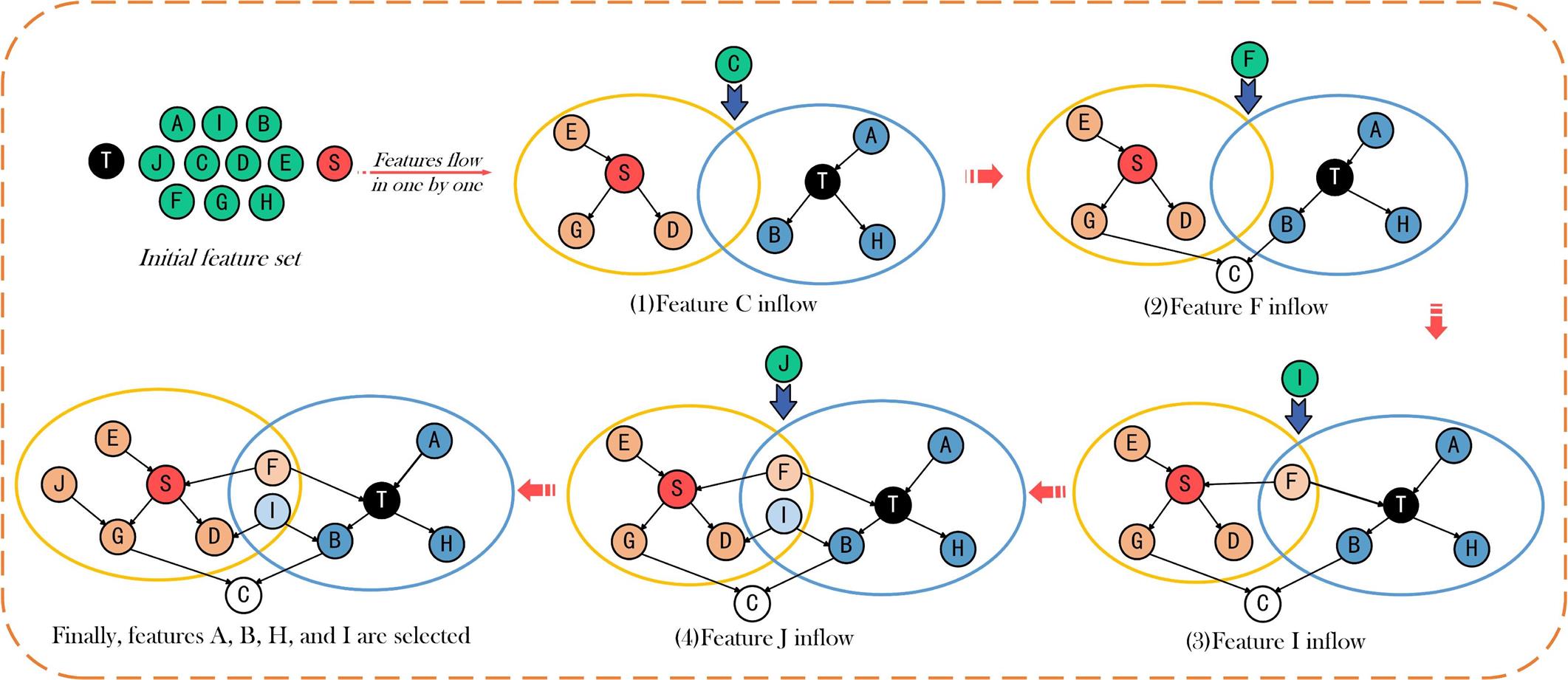}
 \caption{Flowchart of the FairSFS Algorithm}
 \label{fig4-1}
\end{figure}
Figure 2 describes a feature selection algorithm process aimed at handling the relationship between the sensitive feature \(S\) and the target feature \(T\). In this algorithm, features are considered one by one and assigned to sets \(MB_S\) (a set of features related to the sensitive feature \(S\)) and \(MB_T\) (a set of features related to the target feature \(T\)).

Among the features related to the sensitive feature \(S\), there are some features \(X\) that are related to \(S\), i.e., \(X \not\!\perp\!\!\!\perp S \mid MB_{S}\), hence these features \(X\) should not be included in \(MB_T\) to avoid introducing sensitive information into the target feature analysis. However, a dependency path from \(X\) to \(S\) can be blocked by choosing a subset \(Z\) of \(MB_S\) such that \(X\) is conditionally independent of \(S\) given \(Z\) (\(X \perp\!\!\!\perp S \mid Z\)), where \(Z \subseteq MB_{S}\).

Furthermore, by not using the sensitive variable \(S\) itself (which can be viewed as intervening on \(S\) and removing its direct effect on \(T\)), it prevents the transmission of sensitive information through \(X\). According to Lemma 1, if there exists a subset \(Z\) such that \(X \perp\!\!\!\perp S \mid Z\), then \(X\) can be used, as it does not transmit sensitive information to the model.

Therefore, the key step of the algorithm is to identify features incorrectly included in \(MB_S\) from among those related to \(S\) and to find an appropriate conditioning set \(Z\) for these features that effectively blocks the path between these features and the sensitive variable \(S\). Thus, using these screened features to train classifiers can avoid discrimination issues caused by the use of sensitive information. In this way, the algorithm ensures the handling of sensitive features while maintaining accurate identification and analysis of the target features.

\section{Experiments}\par
In this part, we assess the precision and equity of the FairSFS approach over seven fairness-oriented classification datasets, contrasting it with four streaming feature selection methods and two fairness-aware feature selection algorithms. \par

\subsection{Experimental setup}\par
To examine the effectiveness and fairness of the FairSFS approach, experiments were conducted on seven actual datasets, contrasting it with four streamfeature selection algorithms and two fairness-oriented feature selection methods. The comparative analysis involved four methods for streamfeature selection: OSFS, SAOLA, O-DC, OCFSSF; as well as two fairness-aware approaches, Auto and seqsel.

\begin{table}[ht]
\centering
\begin{tabular}{lccc}
\hline
Datasets                   & Samples.num & Features.num & Sensitive feature \\ \hline
Law                        & 20798       & 11           & race              \\
Oulad                      & 21562       & 10           & gender            \\
German                     & 1000        & 20           & age               \\
Compas                     & 6172        & 8            & gender            \\
CreditCardClients          & 30000       & 23           & gender            \\
StudentPerformanceMath     & 395         & 32           & gender            \\
StudentPerformancePort     & 649         & 32           & gender            \\ \hline
\end{tabular}
\caption{Datasets}
\label{tab:your_label}
\end{table}
The significance level for the $G^2$ independence test is set at 0.01. The algorithms are as follows:
\begin{itemize}
\item \textbf{OSFS:} The algorithm detects incoming features via redundancy evaluation and eliminates superfluous characteristics from the chosen set by integrating the freshly introduced features.

\item \textbf{SAOLA:} Carries out redundancy evaluation grounded in information theory and eliminates redundant features throughout sequential assessment.

\item \textbf{SCFSSF:} Continuously recognizes MBs to encapsulate causal links between categorical variables and attributes.

\item \textbf{O-DC:} When new features arrive, O-DC learns PCs and spouses (i.e., MB) conditioned on the currently selected MB through sequential comparison of mutual information within the current PCs, unlike O-ST which learns simultaneously.

\item \textbf{Auto:} The Auto algorithm first trains a classifier for each feature, then selects features with the best AUC metric to combine with the remaining features, and retrains classifiers in subsequent rounds until the end of a 100-round cycle.

\item \textbf{Seqsel:} Seqsel identifies fair features by confirming the independence of attributes from the class variable, conditional on an appropriate set of features, using the Rcit conditional independence test.

\end{itemize}

\textbf{Datasets:} To evaluate the performance of the FairSFS algorithm, we conducted experiments using seven publicly accessible datasets that are frequently utilized for fairness classification tasks. The specifics of these datasets are presented in Table 1. After a comprehensive examination of fair datasets, we meticulously followed established protocols for the management of attribute values, treatment of missing data, and the selection of sensitive features.

\textbf{Classifiers and Evaluation Metrics:} We utilized FairSFS and the comparative algorithms on the aforementioned datasets to derive the features selected by each method. Subsequently, We developed a standardized collection of classifiers—comprising Logistic Regression (LR), Naive Bayes (NB), and k-Nearest Neighbors (KNN)—for each dataset. To gauge the efficacy of these classifiers, we conducted ten-fold cross-validation for each dataset and appraised them using the following performance indicators:

\begin{itemize}

\item \textbf{Accuracy (ACC):} Accuracy refers to the percentage of test samples correctly classified out of all samples. Higher values indicate greater accuracy.

\item \textbf{Statistical Parity Difference (SPD):} SPD measures the extent of disparity in classification outcomes across different groups (frequently based on sensitive attributes like gender or race). This metric is formulated as follows: \begin{equation}
SPD = \mid P(Z' = 1 \mid S = s_{1}) - P(Z' = 1 \mid S = s_{2}) \mid
\end{equation} The SPD ranges from 0 to 1, with lower values indicating a model that exhibits greater fairness.
\item \textbf{Predictive Equality (PE):} This necessitates that the rates of false positives (i.e., the likelihood that a person with a negative outcome is incorrectly predicted as positive) are similar across different groups. This metric is formulated as follows: \begin{equation}
PE = \mid P(Z' = 1 \mid S = 1, Z = 0) - P(Z' = 1 \mid S \neq 1, Z = 0) \mid
\end{equation} The PE ranges from 0 to 1, with lower values indicating a model that exhibits greater fairness.

\end{itemize}

\subsection{ Comparison  with  streaming feature selection}\par
In this section, we compare the FairSFS algorithm with four  streaming feature selection algorithms (OSFS, SAOLA, O-DC, OCFSSF) across seven different datasets. The outcomes, encompassing mean accuracy and fairness measures derived from 10-fold cross-validation, are consolidated in Tables 2, 3, and 4. From this, we can infer the following insights:

\begin{table}[h]
\setlength\tabcolsep{2pt}
\small
\centering
\centering\caption{Comparison of FairSFS, OSFS,SAOLA,O-DC,OCFSSF on KNN Classifier($\uparrow$ indicates that a higher value of the metric is better, while $\downarrow$ indicates that a lower value of the metric is better).}
\label{tab:my-table}
\begin{tabular}{cc||cccccccc}
\hline
metric &Algorithm        & German          & Compas          & Credit          & Law             & Oulad           & Studentm        & Studentp        \\
\hline
\multirow{5}{*}{ACC $\uparrow$}
&OSFS    & 0.6390 &	\textbf{0.5847} &	\textbf{0.7346} 	&0.8042 	&0.5896 	&\textbf{0.9190} 	&0.8982 \\
&SAOLA   & 0.6450 & 0.5624 &    0.2747  &0.8878     &0.6774     &0.9186	&\textbf{0.9106} \\
&OCFSSF  & 0.6410 &	0.5620 &	0.7200  &0.8019 	&0.5878 	&\textbf{0.9190} 	&0.8998 \\
&O-DC     & \textbf{0.6530} &0.5833 	&    0.7222 &0.7855 	&0.6343 	&\textbf{0.9190} 	&0.9075\\
&FairSFS  &0.6090 	&0.5240 	&0.3935 	&\textbf{0.8894} 	&\textbf{0.6784 }	&0.8328 	&0.8705\\
\hline
\multirow{5}{*}{SPD $\downarrow$}
&OSFS    &0.1017 	&0.1795 	&0.0169 	&0.0157 	&0.0171 	&0.1513 	&0.0795\\
&SAOLA   & 0.1329 	&0.0478 	&\textbf{0.0044} 	&0.0479 	&0.0146 	&0.1423 	&0.0804 \\
&OCFSSF  & \textbf{0.0982} 	&0.1212 	&0.0129 	&0.0156 	&0.0141 	&0.1514 	&0.0754 \\
&O-DC     & 0.1222 	&0.1964 	&0.0378 	&0.0096 	&0.0260 	&0.1514 	&0.0793\\
&FairSFS  &0.1082 	&\textbf{0.0239} 	&0.0107 	&\textbf{0.0000} 	&\textbf{0.0073} 	&\textbf{0.1419 }	&\textbf{0.0467}\\
\hline
\multirow{5}{*}{PE $\downarrow$}
&OSFS    &0.1022 	 &0.1621 	&0.0120 	&0.0677 	&0.0278 	&0.0845 	&0.0778 \\
&SAOLA   & 0.1235 	&\textbf{0.0365} 	&\textbf{0.0027} 	&0.0948 	&0.0131 	&\textbf{0.0734} 	&0.0836 \\
&OCFSSF  & \textbf{0.0900} 	&0.1181 	&0.0120 	&0.0782 	&0.0229 	&0.0845 	&0.0771 \\
&O-DC     & 0.1189 	&0.1721 	&0.0389 	&0.0564 	&0.0253 	&0.0845 	&0.0791\\
&FairSFS  &0.1247 	&0.0433 	&0.0157 	&\textbf{0.0000} 	&\textbf{0.0068} 	&0.0820 	&\textbf{0.0365}\\
\hline
\end{tabular}
\end{table}
\begin{table}[!h]
\setlength\tabcolsep{2pt}
\small
\centering
\centering\caption{Comparison of FairSFS, OSFS,SAOLA,O-DC,OCFSSF on NB Classifier.}
\label{tab:my-table}
\begin{tabular}{cc||cccccccc}
\hline
metric &Algorithm        & German          & Compas          & Credit          & Law             & Oulad           & Studentm        & Studentp        \\
\hline
\multirow{5}{*}{ACC $\uparrow$}
&OSFS    &0.6928 	&0.6706 	&0.7755 	&0.8016 	&0.6781 	&0.9189 	&0.9136\\
&SAOLA   &0.6850 	&0.6607 	&\textbf{0.7801} 	&0.8535 	&0.6784 	&\textbf{0.9190} 	&\textbf{0.9168}\\
&OCFSSF  & 0.6990 	&\textbf{0.6732} 	&0.7735 	&0.8219 	&0.6785 	&0.9172 	&0.9152\\
&O-DC     & \textbf{0.7110} 	&0.6719 	&0.7780 	&0.8007 	&0.6785 	&\textbf{0.9190} 	&\textbf{0.9168}\\
&FairSFS  &0.6390    & 0.5709   & 0.7735     & \textbf{0.8761}   & \textbf{0.6786}   & 0.9160   & 0.8736  \\
\hline
\multirow{5}{*}{SPD $\downarrow$}
&OSFS    &0.0929 	&0.2609 	&0.0359 	&0.0302 	&0.0130 	&0.1423 	&0.0937\\
&SAOLA   & 0.1191 	&0.0717 	&0.0213 	&0.0468 	&0.0176 	&0.1423 	&0.0900 \\
&OCFSSF  &\textbf{0.0798} 	&0.1648 	&0.0364 	&0.0226 	&0.0106 	&0.1373 	&0.0894\\
&O-DC     & 0.1222 	&0.1964 	&0.0378 	&0.0096 	&0.0260 	&0.1514 	&\textbf{0.0793}\\
&FairSFS  &0.0986 	&\textbf{0.0180} 	&\textbf{0.0116} 	&\textbf{0.0063} 	&\textbf{0.0084} 	&\textbf{0.1368} 	&0.0850\\
\hline
\multirow{5}{*}{PE $\downarrow$}
&OSFS    &0.1136 	&0.1850 	&0.0247 	&0.0615 	&0.0132 	&0.0734 	&0.0763 \\
&SAOLA   & 0.1104 	&0.0546 	&0.0099 	&0.0949 	&0.0191 	&0.0734 	&0.0780\\
&OCFSSF  &0.0852 	&0.1068 	&0.0270 	&0.0613 	&0.0106 	&\textbf{0.0719} 	&\textbf{0.0710}\\
&O-DC     & \textbf{0.0701 }	&0.2669 	&0.0309 	&0.0381 	&0.0104 	&0.1423 	&0.0905\\
&FairSFS  &0.1389 	&\textbf{0.0463} 	&\textbf{0.0158} 	&\textbf{0.0181 }	&\textbf{0.0089 }	&0.0877 	&0.0901\\
\hline
\end{tabular}
\end{table}

\begin{table}[h]
\setlength\tabcolsep{2pt}
\small
\centering
\centering\caption{Comparison of FairSFS, OSFS,SAOLA,O-DC,OCFSSF on LR Classifier.}
\label{tab:my-table}
\begin{tabular}{cc||cccccccc}
\hline
metric &Algorithm        & German          & Compas          & Credit          & Law             & Oulad           & Studentm        & Studentp        \\
\hline
\multirow{5}{*}{ACC $\uparrow$}
&OSFS    &\textbf{0.7240} 	&0.6733 	&0.8064 	&0.8820 	&0.6858 	&0.9189 	&0.9204\\
&SAOLA   &0.6850 	&0.6607 	&0.7800 	&\textbf{0.8898} 	&0.6797 	&\textbf{0.9190} 	&0.9254\\
&OCFSSF  &0.7230 	&0.6769 	&\textbf{0.8065} 	&0.8824 	&0.6859 	&0.9187 	&\textbf{0.9273}\\
&O-DC     &0.7090 	&\textbf{0.6777} 	&0.8055 	&0.8817 	&\textbf{0.6864} 	&0.9121 	&0.9270\\
&FairSFS  &0.6960 	&0.5710 	&0.7788 	&\textbf{0.8898} 	&0.6785 	&0.9115 	&0.8937  \\
\hline
\multirow{5}{*}{SPD $\downarrow$}
&OSFS    &0.1146 	&0.2134 	&0.0267 	&0.0044 	&0.0145 	&0.1453 	&0.0789\\
&SAOLA   &0.1191 	&0.0717 	&0.0038 	&0.0479 	&0.0126 	&0.1423 	&0.0790 \\
&OCFSSF  &0.1158 	&0.1759 	&0.0268 	&0.0046 	&0.0141 	&\textbf{0.1413} 	&0.0790\\
&O-DC     &\textbf{0.0894} 	&0.2555 	&0.0250 	&0.0050 	&0.0166 	&0.1423 	&0.0790\\
&FairSFS  &0.0116 	&\textbf{0.0180} 	&\textbf{0.0000} 	&\textbf{0.0000} 	&\textbf{0.0031} 	&0.1453 	&\textbf{0.0684}\\
\hline
\multirow{5}{*}{PE $\downarrow$}
&OSFS    &0.1094 	&0.1507 	&0.0147 	&0.0263 	&0.0129 	&0.0734 	&0.0642 \\
&SAOLA   &0.1104 	&0.0546 	&0.0022 	&0.0948 	&0.0206 	&0.0734 	&0.0643\\
&OCFSSF  &0.1078 	&0.1144 	&0.0151 	&0.0251 	&0.0119 	&\textbf{0.0731} 	&0.0643\\
&O-DC    &0.0747 	&0.1790 	&0.0132 	&0.0257 	&0.0133 	&0.0734 	&0.0643\\
&FairSFS  &\textbf{0.0196} 	&\textbf{0.0463} 	&\textbf{0.0000} 	&\textbf{0.0000} 	&\textbf{0.0031} 	&0.0959 	&\textbf{0.0639}\\
\hline
\end{tabular}
\end{table}

\begin{figure}[!h]
\centering
 \includegraphics[width=4.5in, height=1.8in]{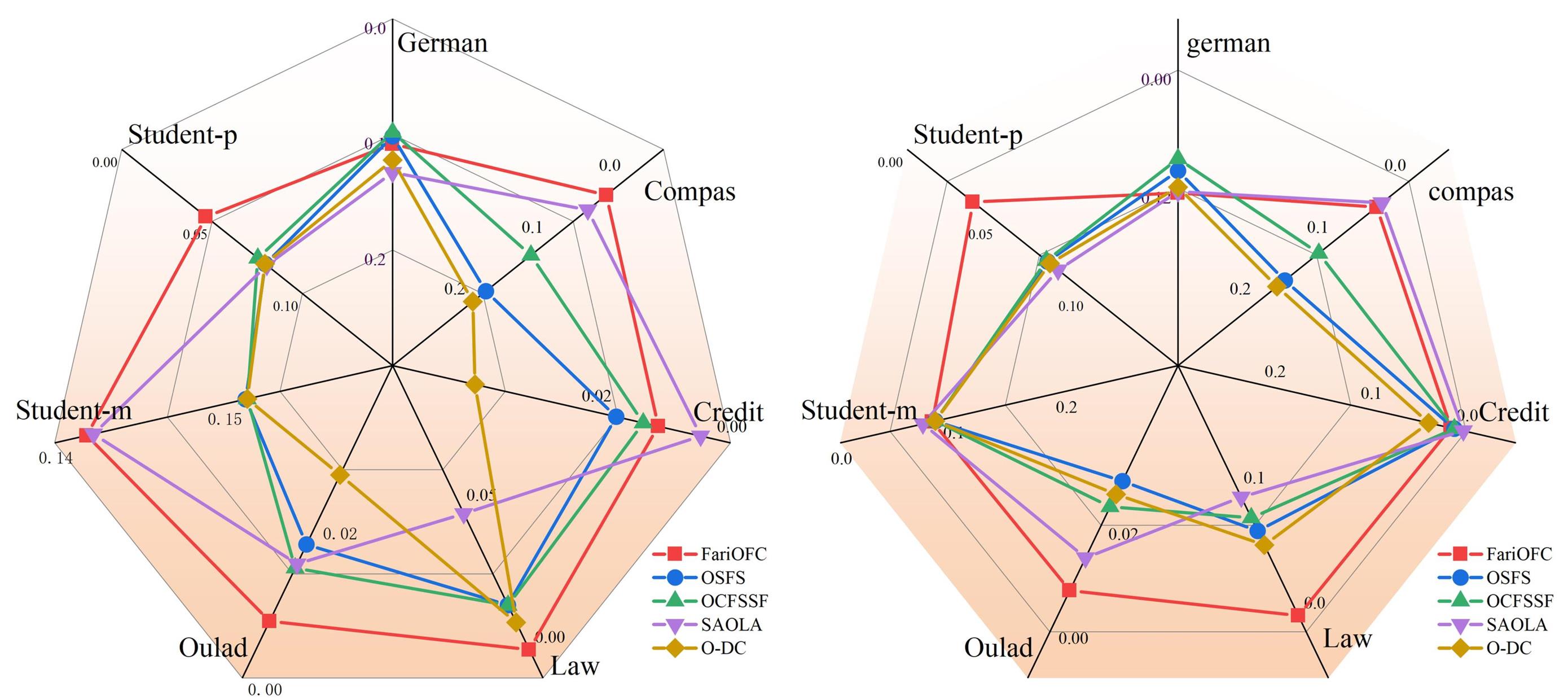}
 \caption{Radar graph depicting the fairness performance of FairSFS alongside its competitors in streaming feature selection, focusing on the metrics SPD (left) and PE (right) when using the KNN classifier(where lower scores for SPD and PE denote increased fairness in the model).}
\end{figure}

\begin{figure}[!h]
\centering
 \includegraphics[width=4.5in, height=1.8in]{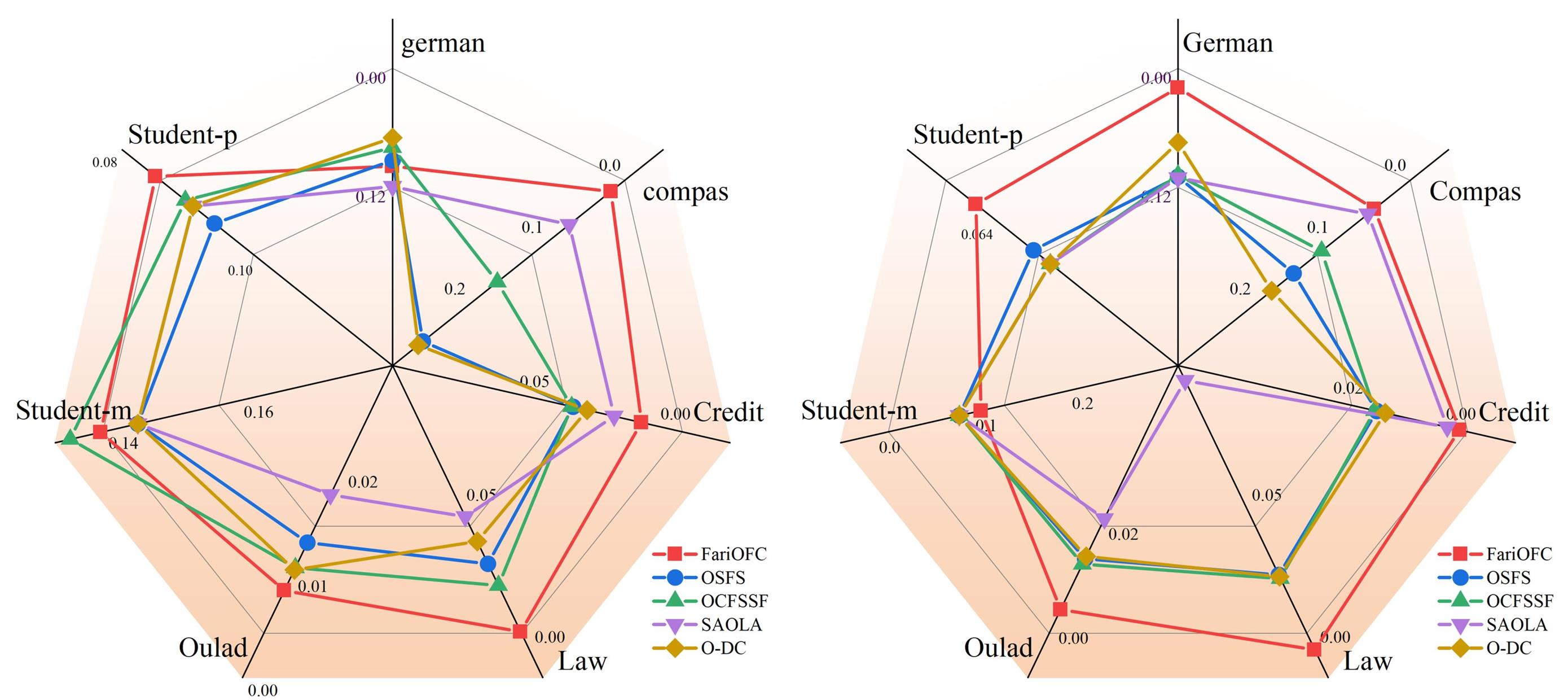}
 \caption{Radar graph depicting the fairness performance of FairSFS alongside its competitors in streaming feature selection, focusing on the metrics SPD (left) and PE (right) when using the NB classifier.}

\end{figure}

\begin{figure}[!h]
\centering
 \includegraphics[width=4.5in, height=1.8in]{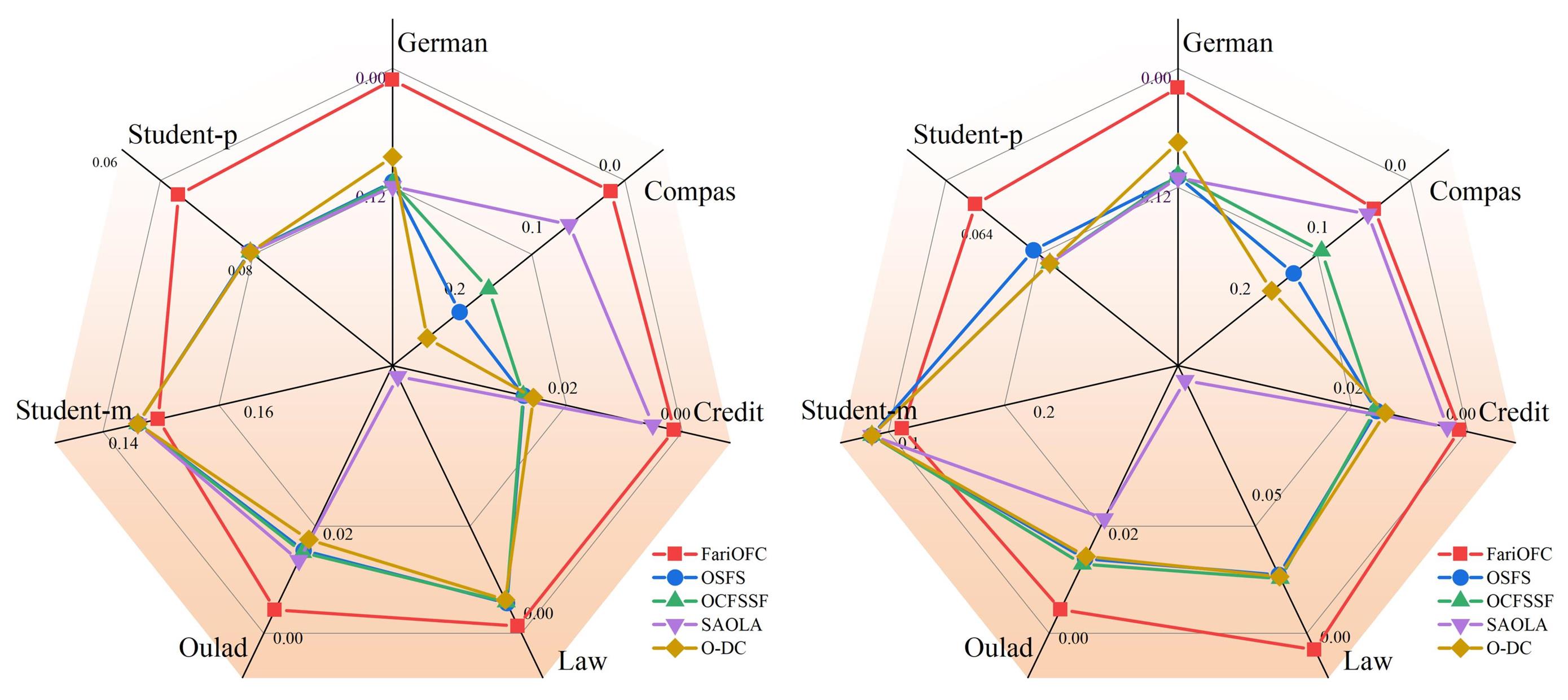}
 \caption{Radar graph depicting the fairness performance of FairSFS alongside its competitors in streaming feature selection, focusing on the metrics SPD (left) and PE (right) when using the LR classifier.}

\end{figure}
\begin{figure}[!h]
\centering
 \includegraphics[width=4.8in, height=1.3in]{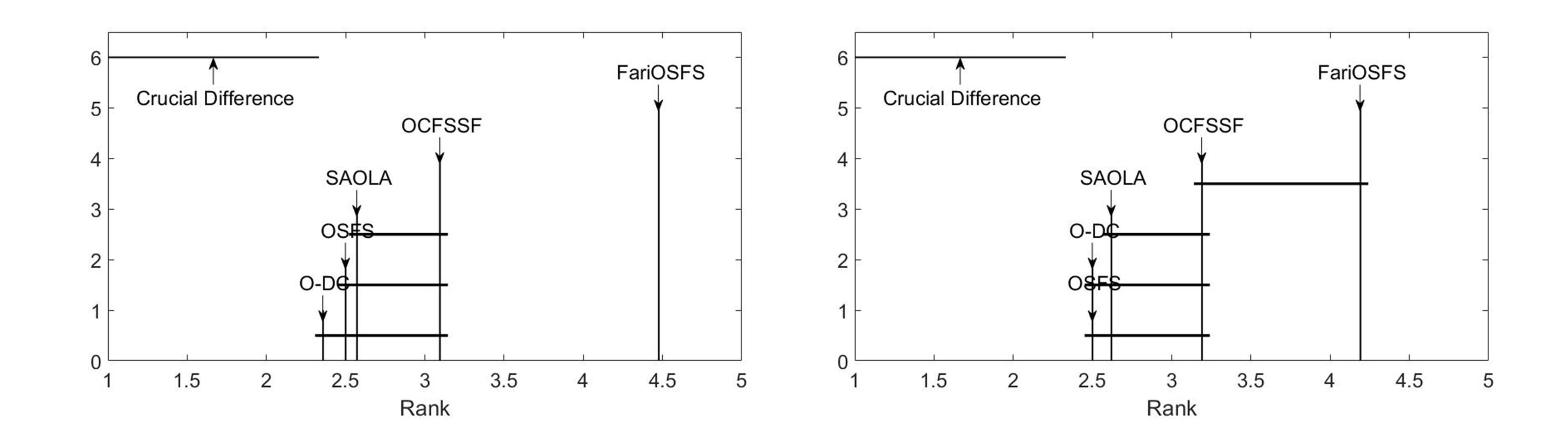}
 \caption{The critical difference plot of the Nemenyi test displays the results of the fairness metric SPD (on the left) and PE (on the right) for FairSFS and its competitors on 7 real-world datasets, with higher rankings indicating better outcomes.}
\end{figure}
\textbf{Accuracy}: The accuracy metrics presented in Tables 2, 3, and 4 reveal that FairSFS attains the highest accuracy on only one or two datasets, with its overall performance generally inferior to that of other algorithms across the majority of the datasets enumerated. Notably, on the German and Compas datasets, FairSFS exhibits a considerably lower accuracy compared to its counterparts. On the remaining datasets, although FairSFS fails to reach the zenith of accuracy, the discrepancy is relatively modest. The FairSFS algorithm, in its quest to eliminate unfair nodes from the Markov Blanket (MB), inherently incurs a trade-off with accuracy. In its pursuit of fairness and real-time performance, FairSFS may sacrifice some degree of accuracy. Nonetheless, the experimental findings suggest that the accuracy of the FairSFS algorithm has not experienced a significant decline and remains competitive with other algorithms.

\textbf{Fairness}: The data presented in Tables 2, 3, and 4 definitively illustrate that although FairSFS achieves accuracy commensurate with other feature selection algorithms, it significantly excels in terms of fairness metrics on the majority of the datasets, outperforming its counterparts by a considerable margin. This outcome attests to the algorithm's efficacy not only in tackling the challenges of streaming feature selection but also in advancing the pursuit of fairness in machine learning models.\par
FairSFS attains peak fairness in three to six datasets across the three classifiers evaluated, with the fairness metrics for the Credit and Law datasets diminishing to negligible levels when using the LR classifier. These two datasets, characterized by their substantial sample sizes, facilitate a more nuanced discernment by FairSFS of the inter relationships among features, class labels, and sensitive variables. This enhanced discernment empowers the conditional independence tests within FairSFS to operate with heightened efficacy, facilitating the identification of fair features across diverse datasets. This deeper level of understanding promotes a more enlightened and judicious feature selection process, thereby exerting a more pronounced influence on fairness metrics. Concurrently, the accuracy of the FairSFS algorithm remains in proximity to that of conventional streaming feature selection algorithms, while concurrently achieving superior fairness outcomes.\par

To visually underscore the fairness advantages of the FairSFS algorithm over four other streaming feature selection methods, we present a comparative line graph. As depicted in Figures 3, 4, and 5, the FairSFS algorithm consistently attains superior fairness metrics across the majority of datasets, with the German, Compas, Credit, Law, and Oulad datasets registering the lowest fairness scores. This signifies that FairSFS is highly effective in purging unfair features during the feature selection process, thereby ensuring a more equitable outcome.

To underscore the fairness advantages of the FairSFS algorithm over other streaming feature selection methods, we performed a Friedman test at a 5\% significance level on the outcomes of three classifiers (SPD and PE). The average rankings for SPD metrics of FairSFS, OSFS, OCFSSF, SAOLA, and O-DC were 4.48, 2.50, 3.10, 2.57, and 2.36, respectively, while the average rankings for PE metrics were 4.19, 2.50, 3.19, 2.62, and 2.50, respectively. The critical difference for FairSFS was 1.33, indicating its significant superiority over the competitors. The critical difference plot for the Nemenyi test is shown in Figure 6.

\subsection{ Comparison with fair feature selection algorithms}\par
In this section, we assess the efficacy of the FairSFS algorithm compared to the Auto and Seqsel algorithms across seven diverse datasets.
 The findings, which consist of mean accuracy and fairness indices obtained from 10-fold cross-validation, are presented in Tables 5, 6, and 7. The following inferences can be made:

\begin{table}[h]
\setlength\tabcolsep{2pt}
\small
\centering
\centering\caption{FairSFS, Auto, and Seqsel are compared using the LR Classifier ($\uparrow$ signifies that a higher metric value is preferable, where as  $\downarrow$  indicates that a lower metric value is more favorable).}
\label{tab:my-table}
\begin{tabular}{cc||cccccccc}
\hline
metric &Algorithm        & German & Compas & Credit & Law    & Oulad  & Studentm & Studentp \\
\hline
\multirow{3}{*}{ACC $\uparrow$} &Auto    & \textbf{0.6990} & \textbf{0.6791} & \textbf{0.7969} & 0.8897 & \textbf{0.6866} & 0.8129   & 0.8689   \\
&Seqsel  & 0.6980 & 0.6296 & 0.7788 & 0.8896 & 0.6808 & 0.8205   & 0.8752   \\
& FairSFS & 0.6960 & 0.5719 & 0.7788 & \textbf{0.8921} & 0.6785 & \textbf{0.9114} & \textbf{0.8937} \\
\hline
\multirow{3}{*}{SPD $\downarrow$}        &Auto    & 0.1156          & 0.2418          & 0.0314          & \textbf{0.0000}          & 0.0194          & 0.1818          & 0.0907          \\
&Seqsel  & 0.0718          & 0.1044          & 0.0000          & 0.0005          & 0.0090          & \textbf{0.0951}          & 0.0935          \\
& FairSFS & \textbf{0.0115} & \textbf{0.0180} & \textbf{0.0000} & \textbf{0.0000} & \textbf{0.0031} & 0.1452 & \textbf{0.0684} \\
\hline
\multirow{3}{*}{PE $\downarrow$} &Auto    & 0.1325          & 0.1702          & 0.0212          & \textbf{0.0000}          & 0.0182          & 0.2965          & 0.0584          \\
&Seqsel  & 0.0569          & 0.0671          & \textbf{0.0000}          & 0.0017          & 0.0099          & 0.3095          & \textbf{0.0381} \\
& FairSFS & \textbf{0.0195} & \textbf{0.0463} & \textbf{0.0000} & \textbf{0.0000} & \textbf{0.0031} & \textbf{0.0959} & 0.0638 \\
\hline
\end{tabular}
\end{table}

\begin{table}[h]
\setlength\tabcolsep{2pt}
\small
\centering
\centering\caption{FairSFS, Auto, and Seqsel are compared using the NB Classifier.}
\label{tab:my-table}
\begin{tabular}{cc||cccccccc}
\hline
metric &Algorithm        & German          & Compas          & Credit          & Law             & Oulad           & Studentm        & Studentp        \\
\hline
\multirow{3}{*}{ACC $\uparrow$} &Auto    & \textbf{0.6840}          & \textbf{0.6800}          & \textbf{0.7970} & 0.6775          & \textbf{0.6790} & 0.7797          & 0.8721          \\
&Seqsel  & 0.6210          & 0.6296          & 0.2935          & \textbf{0.8761}         & 0.6786          & 0.7925          & 0.8582          \\
&FairSFS    & 0.6390          & 0.5709          & 0.7735         & \textbf{0.8761}        & 0.6786      & \textbf{0.9160}          & \textbf{0.8736 }         \\
\hline
\multirow{3}{*}{SPD $\downarrow$}        &Auto    & 0.1450          & 0.2997          & 0.0282          & 0.0326          & 0.0124          & \textbf{0.0924}          & 0.1666          \\
&Seqsel  & 0.1187          & 0.1051          & 0.0136          & 0.0125          & 0.0129          & 0.1176          & 0.1222          \\
&FairSFS   & \textbf{0.0985}&	\textbf{0.0180}	&\textbf{0.0115}	&\textbf{0.0063}	&\textbf{0.0083}	&0.1367	&\textbf{0.0850}         \\
\hline
\multirow{3}{*}{PE $\downarrow$} &Auto    & 0.1656          & 0.2036          & 0.0165          & 0.0480          & 0.0137          & 0.2000          & 0.1124          \\
&Seqsel  & \textbf{0.0875}          & 0.0666          & 0.0191          & 0.0552          & 0.0122          & 0.2121          & 0.0941          \\
&FairSFS     & 0.13889	&\textbf{0.0463}	&\textbf{0.0157}	&\textbf{0.0180}	&\textbf{0.0088}	&\textbf{0.0876}	&\textbf{0.0900}     \\

\hline
\end{tabular}
\end{table}

\begin{table}[h]
\setlength\tabcolsep{2pt}
\small
\centering
\centering\caption{FairSFS, Auto, and Seqsel are compared using the KNN Classifier.}
\label{tab:my-table}
\begin{tabular}{cc||cccccccc}
\hline
metric &Algorithm        & German          & Compas          & Credit          & Law             & Oulad           & Studentm        & Studentp        \\
\hline
\multirow{3}{*}{ACC $\uparrow$} &Auto    & \textbf{0.6660}          & \textbf{0.5928}          & \textbf{0.7884} & 0.7800          & 0.6600          & 0.6987          & 0.8243          \\
&Seqsel  & 0.6470          & 0.5904          & 0.4374          & 0.7093          & 0.6379          & 0.6885          & 0.8351          \\
&FairSFS      & 0.6090	&0.5239&	0.3934	&\textbf{0.8894}	&\textbf{0.6784}	&\textbf{0.8327}	&\textbf{0.8705}          \\
\hline
\multirow{3}{*}{SPD $\downarrow$}        &Auto    & \textbf{0.0855}          & 0.1941          & 0.0507          & 0.0255          & 0.0162          & 0.1470          & 0.0910          \\
&Seqsel  & 0.0999          & 0.1100          & 0.0166          & 0.0198          & 0.0120          & \textbf{0.0992} & 0.0774          \\
&FairSFS          & 0.1082	&\textbf{0.0239}&	\textbf{0.0106}	&\textbf{0.0000}&	\textbf{0.0072}&	0.1418	&\textbf{0.0466}     \\
\hline
\multirow{3}{*}{PE $\downarrow$} &Auto    & \textbf{0.1162}          & 0.1690          & 0.0235          & 0.0726          & 0.0164          & 0.2047          & 0.0961          \\
&Seqsel  & 0.1507          & 0.0936          & 0.0177          & 0.0815          & 0.0130          & 0.3337          & 0.0622          \\
&FairSFS        &  0.1247	&\textbf{0.0433}	&\textbf{0.0157}	&\textbf{0.0000}	&\textbf{0.0068}&	\textbf{0.0820}	&\textbf{0.0365 }      \\
\hline
\end{tabular}
\end{table}

\textbf{Accuracy}: The accuracy scores detailed in Tables 5, 6, and 7 indicate that FairSFS and Auto outperform the competing methods, with each topping the accuracy rankings on 3 to 4 datasets. In contrast, Seqsel’s peak accuracy is limited to a single dataset when paired with the Naive Bayes (NB) classifier. The variability in performance among these algorithms can be attributed to their distinct feature selection approaches. FairSFS and Auto demonstrate strong performance across both accuracy and fairness, while Seqsel’s dedication to fairness may lead to reduced accuracy under certain circumstances. Auto’s feature selection heuristic emphasizes accuracy, selecting features based on their conditional independence from the class and sensitive variables. This method may reveal a greater number of relevant features than FairSFS, thus capturing additional predictive information and yielding higher accuracy.
\begin{figure}[!h]
\centering
 \includegraphics[width=4.5in, height=1.8in]{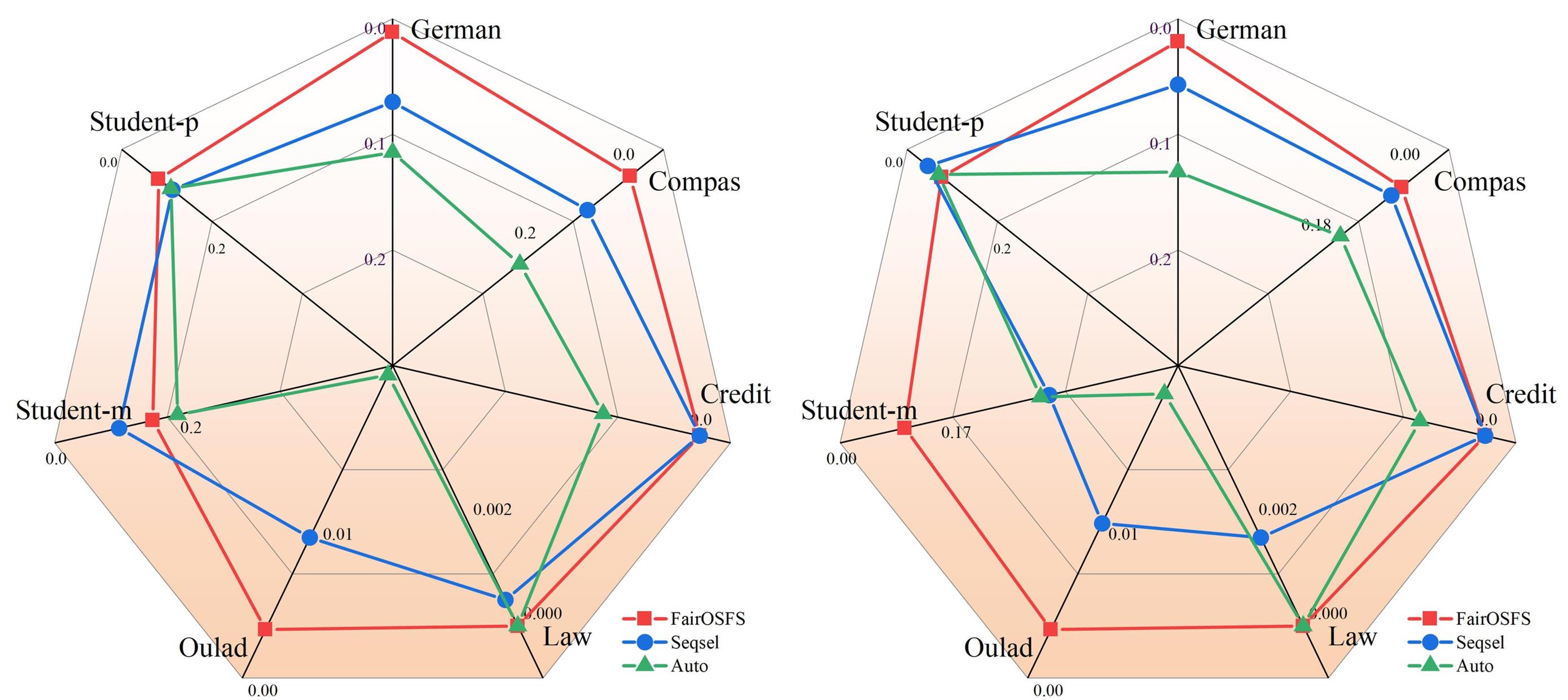}
 \caption{Radar chart comparing the fairness of FairSFS with other methods on SPD (left) and PE (right) metrics using the LR classifier. (where lower scores for SPD and PE denote increased fairness in the model).}

\end{figure}

\begin{figure}[!h]
\centering
 \includegraphics[width=4.5in, height=1.8in]{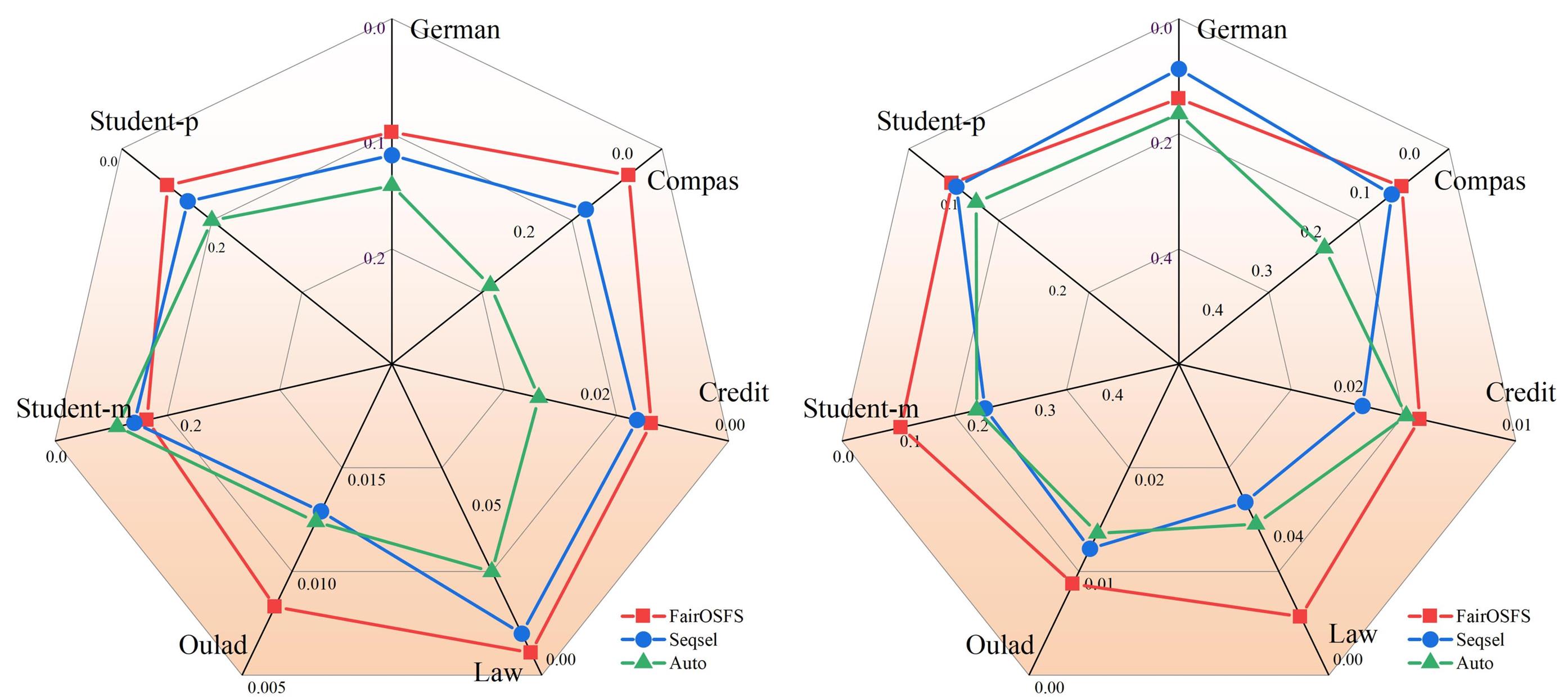}
 \caption{Radar chart comparing the fairness of FairSFS with other methods on SPD (left) and PE (right) metrics using the NB classifier.}

\end{figure}

\begin{figure}[!h]
\centering
 \includegraphics[width=4.5in, height=1.8in]{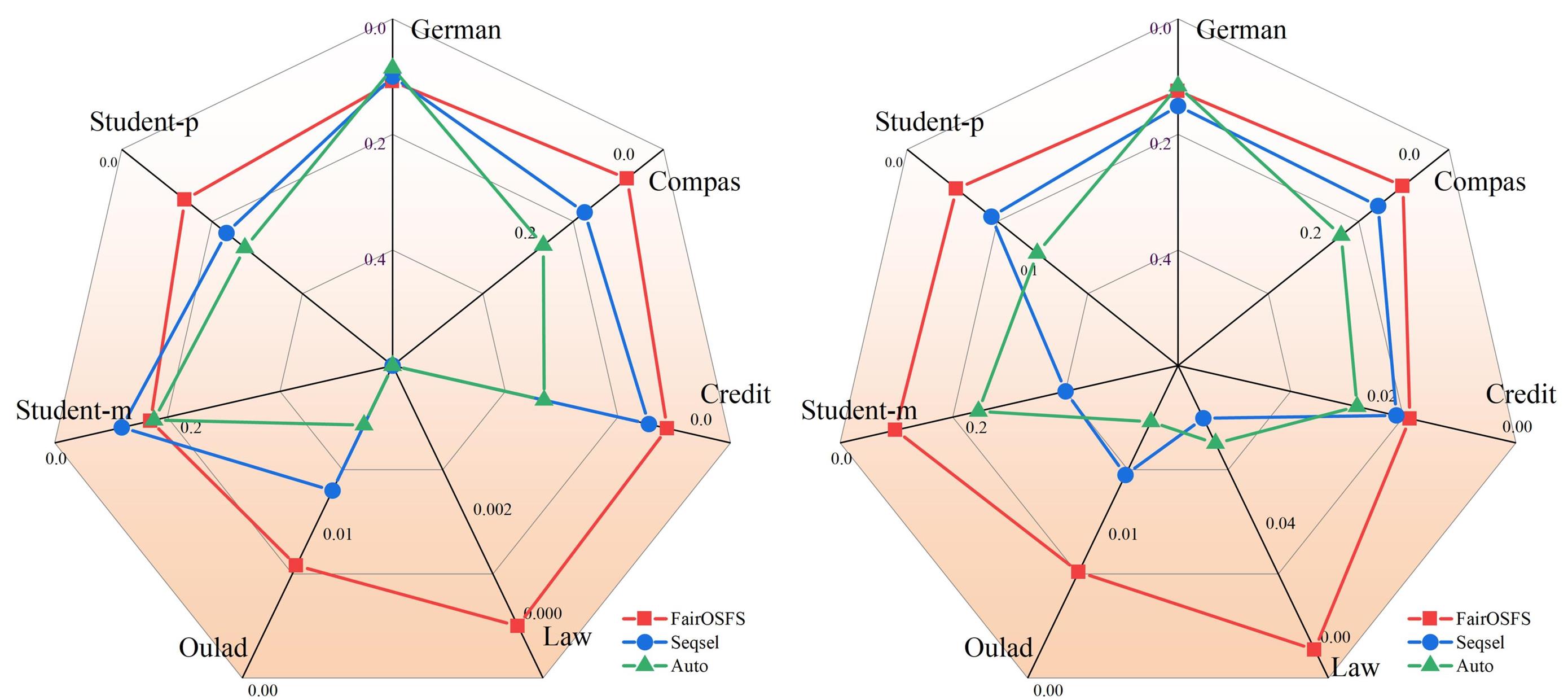}
 \caption{Radar chart comparing the fairness of FairSFS with other methods on SPD (left) and PE (right) metrics using the KNN classifier.}

\end{figure}
\begin{figure}[!h]
\centering
 \includegraphics[width=4.8in, height=1.3in]{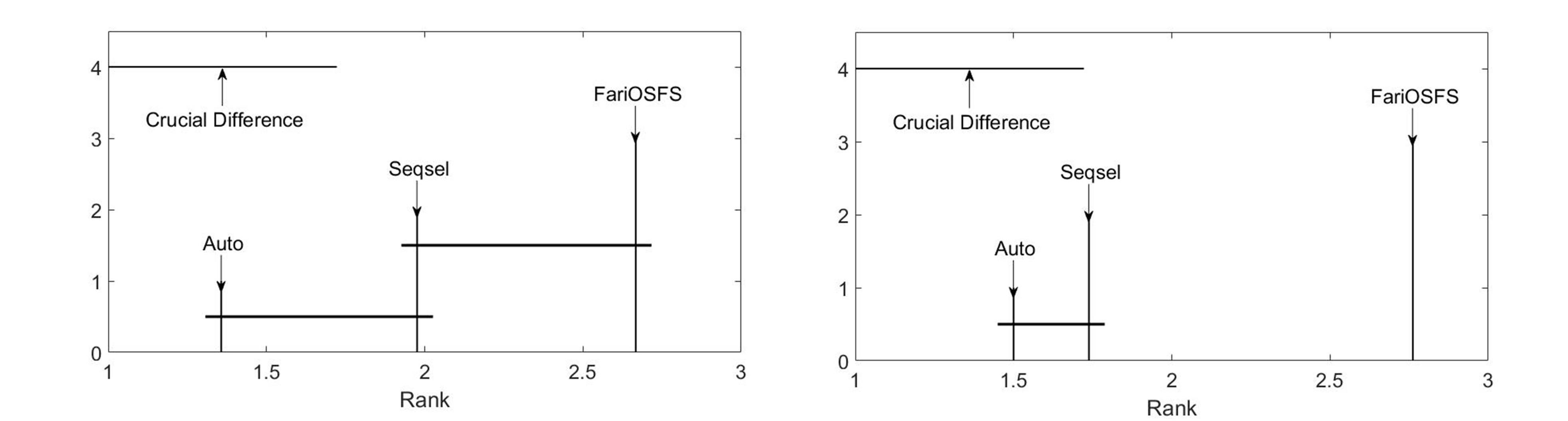}
 \caption{The critical difference plot of the Nemenyi test displays the results of the fairness metric SPD (on the left) and PE (on the right) for FairSFS and its competitors on 7 real-world datasets, with higher rankings indicating better outcomes.}
\end{figure}

\textbf{Fairness}: Tables 5, 6, and 7 clearly show that the FairSFS algorithm achieves better fairness while solving the problem of feature selection stream and maintaining comparable accuracy with other feature selection algorithms.
Figures 7, 8, and 9 demonstrate that FairSFS consistently exhibits the lowest fairness metric across most datasets, particularly excelling in fairness on the german, compas, Credit, Law, and Oulad datasets.To further demonstrate the fairness of the FairSFS algorithm compared to other fair feature selection algorithms, we conducted a Friedman test at a 5\% significance level for the results of three classifiers (SPD and PE). The average rankings for the SPD metric of FairSFS, Seqsel, and Auto were 2.67, 1.98, and 1.36, respectively, while the average rankings for the PE metric were 2.76, 1.74, and 1.50, respectively. The critical difference for FairSFS was 0.72, indicating its significant superiority over the competitors. The critical difference plot for the Nemenyi test is shown in Figure 10.

\section{Conclusion}
Current streaming feature selection algorithms frequently neglect to adequately consider sensitive features within the data, the utilization of which can result in biased and discriminatory model predictions. To rectify this issue, we propose a novel fair stream feature selection algorithm named FairSFS, which can dynamically update the feature set and identify correlations between classification variables and sensitive variables in real time, effectively blocking the flow of sensitive information. The objective of this algorithm is to execute streaming feature selection with a pronounced emphasis on fairness. Experimental evaluations on seven real-world datasets demonstrate that FairSFS exhibits accuracy comparable to other feature selection algorithms, while concurrently addressing the dilemmas of streaming feature selection and attaining enhanced fairness. Nevertheless, it is crucial to acknowledge that with smaller dataset sizes, the $G^2$ test employed by FairSFS for conditional independence assessment may prove inadequate, potentially yielding unforeseen outcomes. Consequently, future inquiries should focus on robustly enhancing fairness in scenarios where dataset sizes are limited.

\bibliographystyle{unsrt}
\bibliography{bare_jrnl}

\end{document}